\documentclass[journal,a4paper]{IEEEtran}
\usepackage[left=1.57cm,right=1.57cm,top=0.95cm,bottom=2.54cm]{geometry}

\usepackage{amsmath,lipsum,calc} 
\usepackage{amssymb}
\usepackage{bm}
\usepackage{acro}
\usepackage{balance}
\usepackage{booktabs}
\usepackage{pifont}
\usepackage{color}
\usepackage[]{graphicx} %
\usepackage[bookmarks=true]{hyperref}
\usepackage{siunitx}    
\usepackage{times}
\usepackage{subcaption}
\usepackage[dvipsnames]{xcolor}
\usepackage{pifont}
\usepackage{enumitem}
\usepackage{multirow, makecell, caption}
\usepackage{rotating, graphicx}
\usepackage{comment}
\usepackage{adjustbox}
\usepackage{tabularx,array}

\usepackage[capitalize]{cleveref}
\crefname{section}{Sec.}{Secs.}
\Crefname{section}{Section}{Sections}
\Crefname{table}{Table}{Tables}
\crefname{table}{Tab.}{Tabs.}
\Crefname{equation}{Equation}{Equations}
\crefname{equation}{Eq.}{Eqs.}

\usepackage[TS1,T1]{fontenc}
\DeclareSIUnit[number-unit-product = {}]{\inch}{\textquotedbl}

\usepackage{tikz,pgfplots}
\usetikzlibrary{matrix}
\pgfplotsset{compat=1.15}
\usepackage{etex} 
\usetikzlibrary{arrows,automata}
\usetikzlibrary{positioning}
\usepgfplotslibrary{groupplots}
\usepackage{tikz-3dplot}
\tikzset{
	state/.style={
		rectangle,
		rounded corners,
		draw=black, very thick,
		minimum height=2em,
		inner sep=2pt,
		text centered,
	},
}
\tikzset{
	info/.style={
		rectangle,
		draw=black, thin,
		minimum height=2em,
		inner sep=2pt,
		text centered,
	},
}
\usetikzlibrary{positioning}
\usetikzlibrary{calc}
\usetikzlibrary{arrows,shapes,backgrounds}
\usepackage{nameref} %

\usepackage[normalem]{ulem}
\newcommand{\change}[1]{#1}
\newcommand{\danping}[1]{#1}
\newcommand{\yunlong}[1]{#1}

\begin{document}
\title{Back to Newton's Laws: Learning Vision-based Agile Flight via Differentiable Physics}

\author{Yuang Zhang$^{1\dag}$, Yu Hu$^{1\dag}$, Yunlong Song$^{\dag}$, Danping Zou$^{1\ast}$, Weiyao Lin$^{1\ast}$\\% <-this %
\normalsize{$^{1}$Shanghai Jiao Tong University, Shanghai, China}\\
\normalsize{$^{\dag}$Equal contribution}
\normalsize{$^\ast$Corresponding  Author}}

\maketitle

\begin{abstract}
Swarm navigation in cluttered environments is a grand challenge in robotics. This work combines deep learning with first-principle physics through differentiable simulation to enable autonomous navigation of multiple aerial robots through complex environments at high speed. Our approach optimizes a neural network control policy directly by backpropagating loss gradients through the robot simulation using a simple point-mass physics model and a depth rendering engine. Despite this simplicity, our method excels in challenging tasks for both multi-agent and single-agent applications with zero-shot sim-to-real transfer. In multi-agent scenarios, our system demonstrates self-organized behavior, enabling autonomous coordination without communication or centralized planning---an achievement not seen in existing traditional or learning-based methods. In single-agent scenarios, our system achieves a 90\% success rate in navigating through complex environments, significantly surpassing the 60\% success rate of the previous state-of-the-art approach.
Our system can operate without state estimation and adapt to dynamic obstacles. In real-world forest environments, it navigates at speeds up to 20 m/s, doubling the speed of previous imitation learning-based solutions. Notably, all these capabilities are deployed on a budget-friendly \$21 computer, costing less than 5\% of a GPU-equipped board used in existing systems.
Video demonstrations are available at \url{https://youtu.be/LKg9hJqc2cc}.

\end{abstract}


\IEEEpeerreviewmaketitle

\section{Introduction}

Modern aerial robotics are increasingly being called upon to execute complex, agile maneuvers in dynamically changing and unknown environments, ranging from search-and-rescue operations \cite{schedl_autonomous_2021} to powerline inspections \cite{xing_autonomous_2023} or deliveries \cite{sage_testing_2022}. 
They are required to operate with extremely limited sensory and computational resources onboard. 
The key to overcoming this challenge is tightly integrating efficient, robust perception systems with adaptive control mechanisms. 
However, existing methods~\cite{giusti2015machine, gao_teach-repeat-replan_2020, zhou_ego-planner_2020, zhou_swarm_2022, loquercio_learning_2021, zhang_perception-aware_2018} for this problem still lacking in terms of robustness, flexibility, and scalability. 
Achieving agile vision-based flight in complex environments with extremely limited sensory and computational resources remains an open research problem in robotics.

Classical approaches predominantly rely on mapping-based methods for navigation\cite{gao_teach-repeat-replan_2020, zhou_ego-planner_2020}.
These methods typically divide the navigation task into four separate stages: localization, mapping, planning, and control. 
Such methods have achieved impressive results in research labs and real-world applications, from exploring Mars to swarm navigation in dense forests~\cite{zhou_swarm_2022}. However, the cascade structure introduces latency and error accumulation, limiting their scalability to high-speed flight \cite{loquercio_learning_2021, zhang_perception-aware_2018}. Furthermore, localization and mapping can be erratic at high speeds and suffer from high computational costs \cite{delmerico2019we, wang_tartanair_2020, teed_droid-slam_2021}.

In contrast to the traditional pipeline, recent studies propose to learn end-to-end policies directly from data without explicit mapping and planning stages. 
Particularly, reinforcement learning (RL) has achieved impressive results in solving complex robot control problems, from quadrupedal locomotion over challenging terrain~\cite{hwangbo2019learning, Miki2022ScienceRobotics, choi2023learning} to reaching the limit in autonomous drone racing~\cite{yunlong_science}. 
However, RL suffers from slow convergence and tends to be data-intensive, often necessitating large-scale, parallelized environments~\cite{makoviychuk2021isaac, akkaya2019solving} for fast data collection and stable policy gradient estimation. 
Consequently, their applicability has largely been confined to state-based control problems, with limited success in tasks that require vision-based inputs.

As an alternative strategy, imitation learning has been widely used in vision-based applications due to its straightforward implementation and high sample efficiency. For instance, a study by \cite{kaufmann_deep_2020} proposes to learn a sensorimotor policy that enables an autonomous quadrotor to fly acrobatic maneuvers by leveraging demonstrations from an optimal controller. 
Similarly, ~\cite{loquercio_learning_2021} proposes using imitation learning to train a neural network that can directly map noisy depth images into collision-free trajectories~\cite{loquercio_learning_2021} for obstacle avoidance. 
This methodology substantially cuts down on inference time, thereby enhancing the system to navigate obstacles and achieve high-speed flight in unstructured environments swiftly. 

Although the imitation learning framework can effectively solve a specific task and is efficient for high-dimensional inputs, they often rely heavily on the quality and comprehensiveness of expert demonstrations. This dependency restricts the system's generalizability beyond the initial training data and can lead to inaccurate imitation of the expert. Additionally, experts are typically tailored for a specific task, making it challenging to solve new tasks. The complex labeling and training processes further restrict its flexibility and limit practical applications.
Therefore, we ask: \emph{how can we optimize a vision-based policy in a scalable and efficient manner while maintaining task flexibility?}

\begin{figure*}
    \centering
    \includegraphics[width=\linewidth]{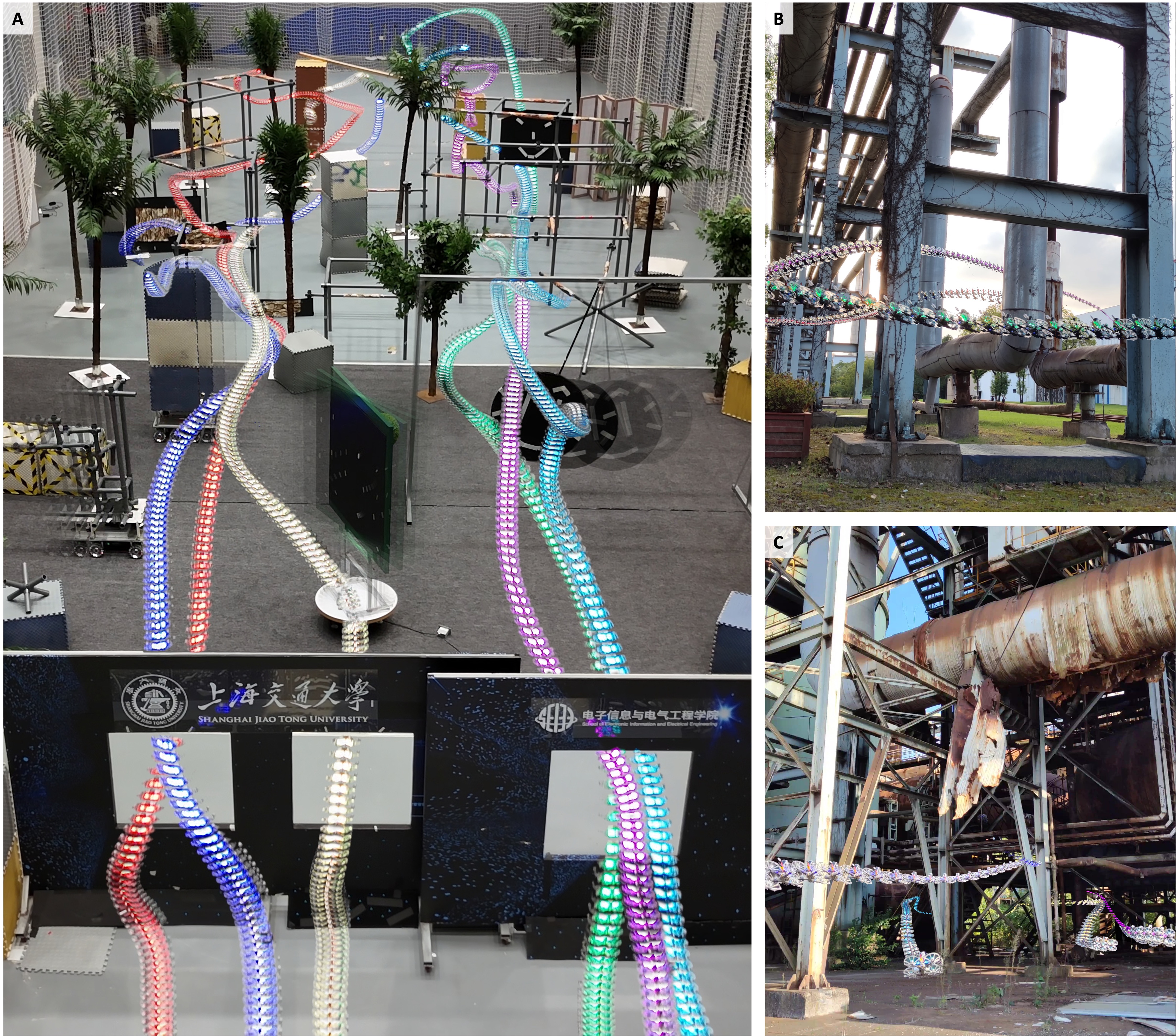}
    \caption{{\bf Vision-based agile swarm navigation through cluttered environments using an end-to-end neural network controller trained via differentiable physics.} (A) The indoor environment features both static and dynamic obstacles. (B,C) Outdoor scenarios. Our system operates on a budget-friendly $\$$21 ARM-based computer coupled with a depth camera for onboard sensing. \danping{All results were achieved using onboard vision sensors without communication between robots.}
    }
    \label{fig:results_overview}
\end{figure*}

\begin{figure*}
    \centering
    \includegraphics[width=\linewidth]{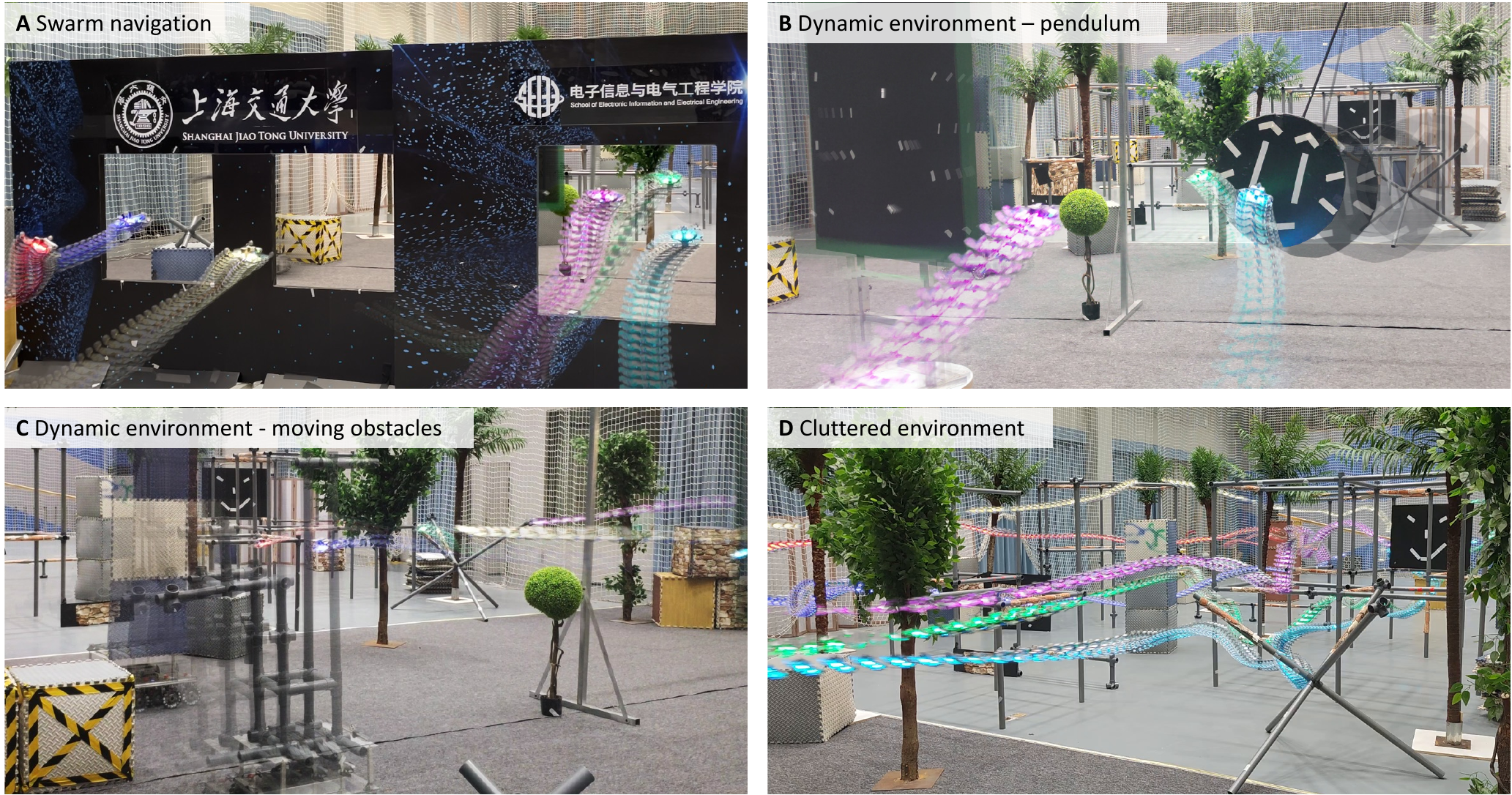}
    \caption{{\bf Detailed perspectives on swarm navigation through a cluttered environment. } 
    A team of six drones flying through gates (A), avoiding dynamic obstacles (B, C), and navigating through thin structures~(D).
    }
    \label{fig:results_enlarged}
\end{figure*}

We explore a different robot learning paradigm by training neural network policies using \emph{differentiable physics.} Specifically, we leverage prior knowledge about the physical system and its inherent differentiability to optimize a deep neural network policy end-to-end.
Differentiable dynamics enable performing backpropagation through a physical simulator directly. This stands in contrast to model-free reinforcement learning and imitation learning; both methods treat the system dynamics as a black box and estimate the policy gradient using only sampled trajectory rollouts. 
We discover that using physics priors, particularly simple point-mass dynamics, for policy optimization enhances the training efficiency, task scalability, and generalization capability of the learned flight system.  

This work demonstrates the effectiveness of \danping{leveraging} differentiable physics in training a robust end-to-end navigation policy for agile vision-based flight in dynamic and unstructured environments.
Our key insight is that the autonomous flight task can be formulated via a physics-driven loss function, which enables the direct computation of policy gradients through backpropagation.
The physics-driven loss function specifies proximity to a velocity target, distance to obstacles, and control smoothness; it can be calculated by rolling out the system dynamics using a neural network control policy and a differentiable simulator. \danping{Both the loss and the dynamics are differentiable.} 
Afterward, the neural network policy is updated via gradient descent, in which the gradient is computed via backpropagation through both the physics simulation and the loss function. 
Training directly with the physics-driven objective function is highly efficient and can be easily scaled up to solve diverse tasks.

\subsection*{Contribution}
We demonstrated the effectiveness of our method by training a recurrent neural network policy for vision-based agile navigation \danping{in both single-agent and multiple-agent applications.} \danping{The results show that the trained policy} can navigate drones at high speed through complex environments featuring both static and dynamic obstacles. Furthermore, we showcased proficiency in agile swarm navigation without the need for communication, as well as flights without dependency on a stand alone odometry module.
In contrast to \danping{traditional mapping-and-planning} approaches, our end-to-end \danping{flight} policy offers substantial advantages in reducing both computational latency and error accumulation. %
\danping{Compared to existing learning-based techniques} ~\cite{giusti2015machine, loquercio_learning_2021, song2023learning, sadeghi2017CAD2RL}, which are predominantly limited to single-agent configurations, often due to the need for expert demonstrations or discretized control spaces, our approach showcases exceptional flexibility and generalization capabilities for more complex, multi-agent scenarios. 

Fig.~\ref{fig:results_overview} provides an overview of our real-world experiments, where our \danping{flight} policy demonstrates reliable high-speed flight in unstructured, dynamic environments.
Fig.~\ref{fig:results_enlarged} shows detailed perspectives on swarm navigation through a cluttered indoor environment.
Although the policy is solely trained in simple simulation environments featuring randomly placed obstacles, it exhibits strong generalization in real-world scenarios, achieving speeds of up to~\SI{20}{\meter\per\second} in the wild. 
\danping{In comparison, the state-of-art method \cite{loquercio_learning_2021} achieves a maximum speed of~\SI{10}{\meter\per\second}. The performance gain in agility is attributed to our physics-informed learning approach and to the flexibility of our end-to-end flight system design. 
Furthermore, our system can avoid moving obstacles without requiring explicit object detection or motion prediction, which significantly enhances the adaptability of our approach in dynamic environments.
} 

Importantly, our physics-driven policy optimization allows for straightforward scaling to multi-agent contexts. After training in a multi-agent setting, the policy (deployed for decentralized control) can navigate multiple drones through narrow gates, while maintaining awareness of other vehicles without communication or global planning \danping{(see Fig. \ref{fig:multi_agent})}. %
Interestingly, the agents exhibit self-organized behaviors such as waiting, following, resolving conflicts, and mutual collision avoidance.
These behaviors emerged from minimizing a simple physics-driven function for obstacle avoidance \danping{without explicitly designing a collaboration reward. Such emerging behaviors have not been demonstrated before in the literature and could inspire further research in this area.}

Finally, \danping{we demonstrate that the trained model achieves} exceptional computational efficiency in real-world deployments. \danping{It operates} on a budget-friendly \$21 ARM-based computer coupled with a depth camera.
This is a crucial advantage over other state-of-the-art systems~\cite{zhou_swarm_2022, loquercio_learning_2021, agilicious, kaufmann_deep_2020} that \danping{rely on} expensive and power-hungry GPUs. While those systems may offer high computational power, they come with prohibitive costs and energy requirements, limiting their applicability in resource-constrained environments. 
\yunlong{This achievement is attributed to the efficiency and effectiveness of our training approach, which allows for the use of simple network architectures and low-resolution inputs to develop robust flight policies. As shown in Fig.~\ref{fig:robustness_sim}E, with the same training setup, our method requires only 10\% of the samples needed by traditional RL methods to reach its maximum reward level.}

\section{Method}

\subsection{Problem Formulation}
Without loss of generality, we formulate the robot navigation task in an optimization framework. The robot is modeled as a discrete-time dynamical system, characterized by continuous state and control input spaces, denoted as \( \mathcal{X}  \) and \( \mathcal{U} \), respectively. At each time step \( k \), the system state is \( x_k \in \mathcal{X} \), and the corresponding control input is \( u_k \in \mathcal{U} \).
An observation \( o_k \in \mathcal{O} \) is generated at each time step based on the current state \( x_k \) through a sensor model \( h: \mathcal{X} \rightarrow \mathcal{O} \), such that \( o_k = h(x_k) \).
The system's dynamics are governed by the function \( f: \mathcal{X} \times \mathcal{U} \rightarrow \mathcal{X} \), which describes the time-discretized evolution of the system as \( x_{k+1} = x_k + f(x_k, u_k) \). The discrete-time instants are \( t_k = \Delta t \cdot k \), where \( k \) ranges from 0 to \( N \), thereby establishing a finite time horizon for the control problem.
Crucially, at each time step \( k \), the robot receives a cost signal \( l_k = l(x_{k}, u_k)\), which is a differentiable function of the current state \( x_{k} \) and the control input \( u_k \). 

We model the policy as a function approximator represented by a neural network~\( u_k = \pi_{\theta} (o_k)\). 
The neural network takes the observation \( o_k \) as input and outputs the control input \( u_k \).
The objective of our optimization problem is to find the optimal policy parameters \(\theta^{\ast}\) by minimizing the total loss via gradient descent:
\begin{align}
    \min_{\theta} \mathcal{L}_{\theta} &=  
      \sum_{k=0}^{N-1} l(x_{k}, u_k) 
    = 
      \sum_{k=0}^{N-1} l(x_{k}, \pi_{\theta}(o_k))
      \\
    \theta & \leftarrow \theta - \gamma \nabla_{\theta} \mathcal{L}_\theta
\end{align}
where $\gamma$ is the learning rate.

\textbf{Policy gradient via differentiable simulation}:
In differentiable simulation for policy learning, the backward pass is crucial for computing the analytic gradient of the objective function with respect to the policy parameters. 
Following \cite{metz_gradients_2021}, the policy gradient can be computed directly via
$\partial \mathcal{L}_{\theta} / \partial \theta = $
\begin{equation*}
      \frac{1}{N}\sum_{k=0}^{N-1} \left( \sum_{i=0}^{k} \frac{\partial l_k }{\partial x_{k}} \prod_{j=i+1}^{k} \left( \frac{\partial x_{j}}{\partial x_{j-1}} \right) \frac{\partial x_i}{\partial \theta} + \frac{ \partial l_k}{\partial u_k} \frac{ \partial u_k}{\partial \theta} \right) 
      ,
\end{equation*}
where the matrix of partial derivatives \( \partial x_j / \partial x_{j-1}\) is the Jacobian of the dynamical system~$f$. 
Therefore, we can compute the policy gradient directly by backpropagating through the differentiable physics model and a physics-driven objective function~\(l_k\) that is differentiable with respect to the physics model. 
This enables more efficient and potentially more accurate optimization compared to traditional reinforcement learning methods.

\subsection{Differentiable physics simulation}
The fundamental difference between a differentiable physics simulation and a standard simulation, e.g. MuJoCo~\cite{todorov2012mujoco} or Flightmare~\cite{song_flightmare_2021}, lies in the ability to compute gradients with respect to model parameters or inputs. 
Such a simulation can propagate the gradient from the output state \( x_k \) directly to the control input $u_k$, and hence, to the policy parameters $\theta$. 

We provide a differentiable physics simulation for quadrotors. 
Typically, a quadrotor is modeled as a rigid body, which is actuated by four motors~\cite{song_flightmare_2021}. 
\danping{Instead of seeking to accurately model the real-world quadrotor dynamics, we opt for using a simple point mass model to approximate the quadrotor dynamics. While common belief suggests that high-fidelity models are necessary for successful simulation-to-reality transfer, our approach demonstrates that this simple model can achieve this goal. Additionally, this approximation enables faster gradient computation \change{and better optimization landscape} for training, significantly improving efficiency.}
To this end, we have based our differentiable physics simulation on a simple point mass model, which is represented by the numerical integration of acceleration and velocity:
\begin{equation*}
  {v}_{k+1}= {v}_k+\frac{{a}_k+{a}_{k+1}}{2} \Delta t, \qquad {p}_{k+1}={p}_k+{v}_k+\frac{1}{2}{a}_k \Delta t^2.
\end{equation*}
\danping{We control the vehicle using a thrust controller, which is tracked by a classic inner-loop flight controller operating at a high frequency.}
To model flight controller response, we implement a fixed control latency and an exponential moving average on the sequence of desired thrust acceleration. This process, as illustrated in Fig.~\ref{fig:method_overview}, can be represented as the convolution between the desired thrust vector \danping{$u_k$} with the simplified flight controller impulsive response:
\begin{equation*}
  \eta(t)=
  \begin{cases}
    0, & t < \tau \\
    \lambda e^{-\lambda (t - \tau)}, & t \geq \tau \\
  \end{cases}
\end{equation*}
where $\tau$ is a fixed delay and $\lambda$ is the exponential smoothing factor.
The filtered thrust \danping{$\hat{u}_k$} is considered as the actual thrust applied on the point mass model. 

We also add a quadratic air drag and take the sum of thrust and air drag to calculate the acceleration (denoted as $a_k$) for the point mass physics simulation. The parameters $\tau$ and $\lambda$ as well as the linear air drag coefficient are tuned to match the real platform through calibration (Supplementary S1).
Since a quadrotor can only generate a collective thrust in its body's upward direction, the magnitude (length) of the thrusts defines the overall vehicle acceleration, while the direction of the thrust vector defines the inclination. 
Additionally, we align the quadrotor's yaw angle with the target direction, establishing a unique orientation. After that, the quadrotor's orientation and position (from the point-mass model) serve as the extrinsic parameters for rendering the depth sensor data.

\subsection{Training environment and physics-driven objective function}

\begin{figure*}[!htp]
    \centering
    \includegraphics[width=\linewidth]{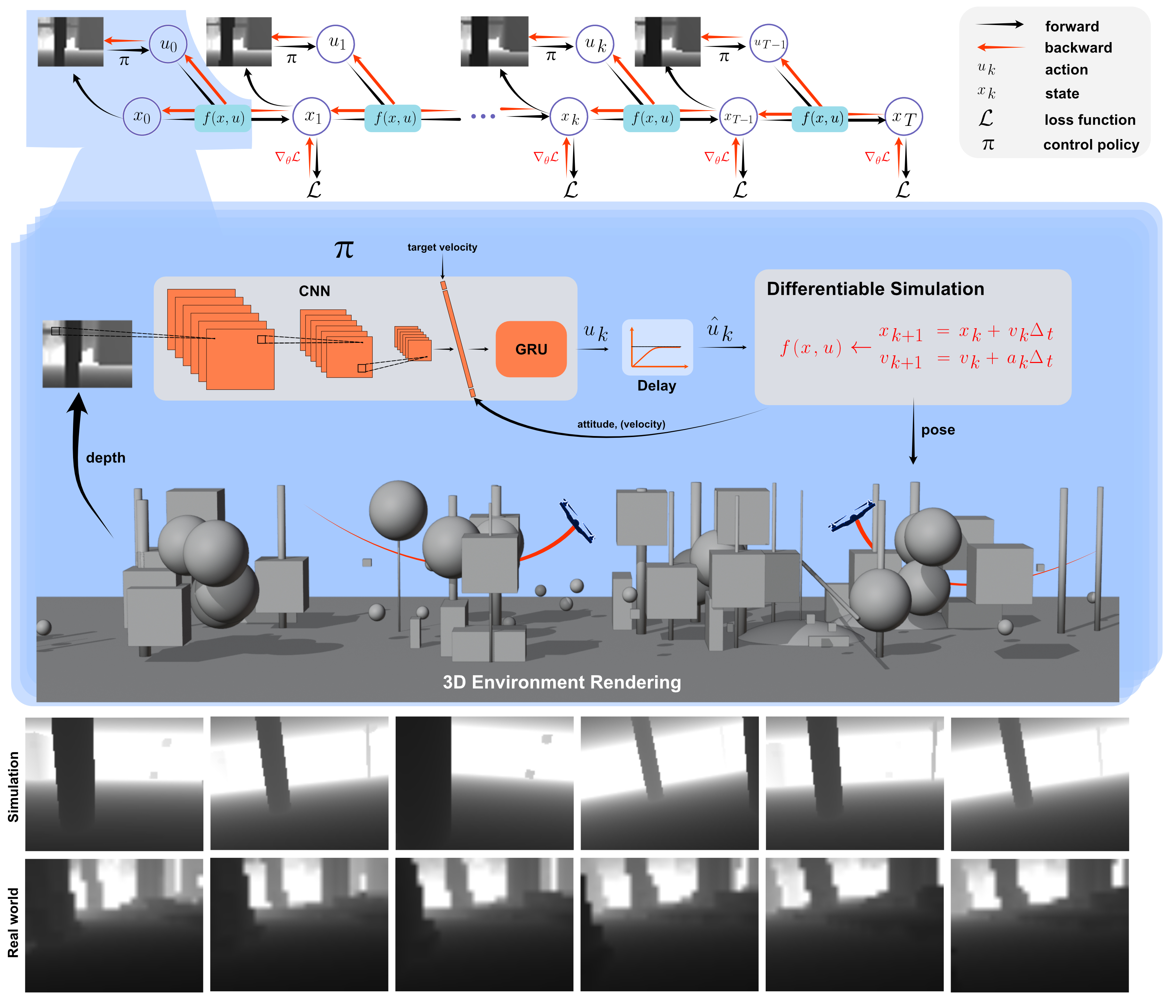}
    \caption{{\bf Method overview: learning vision-based agile flight via differentiable physics.} The differentiable simulator carries gradients from the output state \( x_k \) directly to the control inputs $u_k$, and hence, to the policy parameters $\theta$. A single timestep consists of depth map rendering, action prediction, and quadrotor dynamics simulation. The training environment only includes randomly placed obstacles.}
    \label{fig:method_overview}
\end{figure*}

Our training environment, including depth rendering and physics simulation, is implemented with CUDA as a PyTorch extension \cite{paszke_pytorch_2019}. 
As illustrated in Figure \ref{fig:method_overview}, the training environment features four kinds of abstract obstacles, including planes, cuboids, spheres, and cylinders. \danping{While existing methods for training vision-based policies typically rely on high-fidelity simulators to generate realistic scenes, we discovered that our training approach, even when using such a simple simulation environment, can produce robust vision-based flight policies that generalize well to diverse real-world scenarios.}
We randomly generate the dimensions and positions of the obstacles for each rollout.
There are 64 environments running in parallel for one batch in training. 
At each training iteration, we collect 64 trajectories by simulating the environments for 150 timesteps, with a time step  of 1/15~\SI{}{\second}.

The physics-driven objective function (or loss) consists of three parts: proximity to velocity targets, obstacle avoidance, and control smoothness. 
Proximity to velocity targets is measured using the smooth L1 loss \cite{girshick_fast_2015} between velocity control error and zero,  

\begin{equation*}
    \mathcal{L}_v=\frac{1}{T} \sum_{k=1}^T \text{Smooth L1}(\|v_k^{set}-\bar{v}_k \|_2,0)
\end{equation*}
where $\bar{v}_k$ is the average velocity calculated via moving average spanned across a 2-second time window. 
The average operation enables the agent to focus on obstacle avoidance locally while tracking the desired velocity $v^{set}_k$ in a longer horizon.
The direction of the desired velocity, denoted as $v^{set}_k$, is determined by the difference between the target position and the current position. Its magnitude is limited to a predefined maximum speed, and its gradient is not propagated during training.

The obstacle avoidance loss is defined by the distance between the quadrotor center and the nearest point in the environment. 
We denote the closest distance as $d_k$, the radius of the drone as $r_q$, and the speed approaching the closest point as $v_k^c$. 
The obstacle-avoidance loss is the approaching speed multiple with the sum of a truncated quadratic function and a soft plus barrier:
\begin{equation*}
  \mathcal{L}_c=\frac{1}{T} \sum_{k=1}^T v_k^c \max(1-(d_k-r_q),0)^2+\beta_1 \ln(1+e^{\beta_2 (d_k-r_q)}),
\end{equation*}
where $\beta_1$ and $\beta_2$ are control parameters of the soft plus barrier function. we choose $\beta_1=2.5$ and $\beta_2=32$ for our implementation. The approaching speed $v_t^c$ adjusts the penalty according to how fast the drone is approaching an obstacle. When the drone is stationary or moving away from the obstacle, this factor will nullify the penalty. Conversely, when the drone is moving towards the obstacle, it will ensure that the obstacle avoidance loss is significant enough to slow down the agent.

To avoid large oscillation in the action space, we define a smoothness loss, which is the penalty on the acceleration and jerk:
\begin{equation*}
  \mathcal{L}_a=\frac{1}{T} \sum_{k=1}^T \|a_k \|^2,
  \qquad
  \mathcal{L}_j=\frac{1}{T-1} \sum_{k=1}^{T-1} \|\frac{a_k-a_{k+1}}{\Delta t}\|^2.
\end{equation*}
The final loss is the weighted sum of these four terms, i.e., $L=\lambda_v \mathcal{L}_v+\lambda_c \mathcal{L}_c+\lambda_a \mathcal{L}_a+\lambda_j \mathcal{L}_j$ where $\lambda_v$, $\lambda_c$, $\lambda_a$, and $\lambda_j$ are the weights. We choose $\lambda_v=1, \lambda_c=2, \lambda_a=0.01, \lambda_j=0.001$ for our implementation.

\begin{figure*}[ht!]
    \centering
    \includegraphics[width=\linewidth]{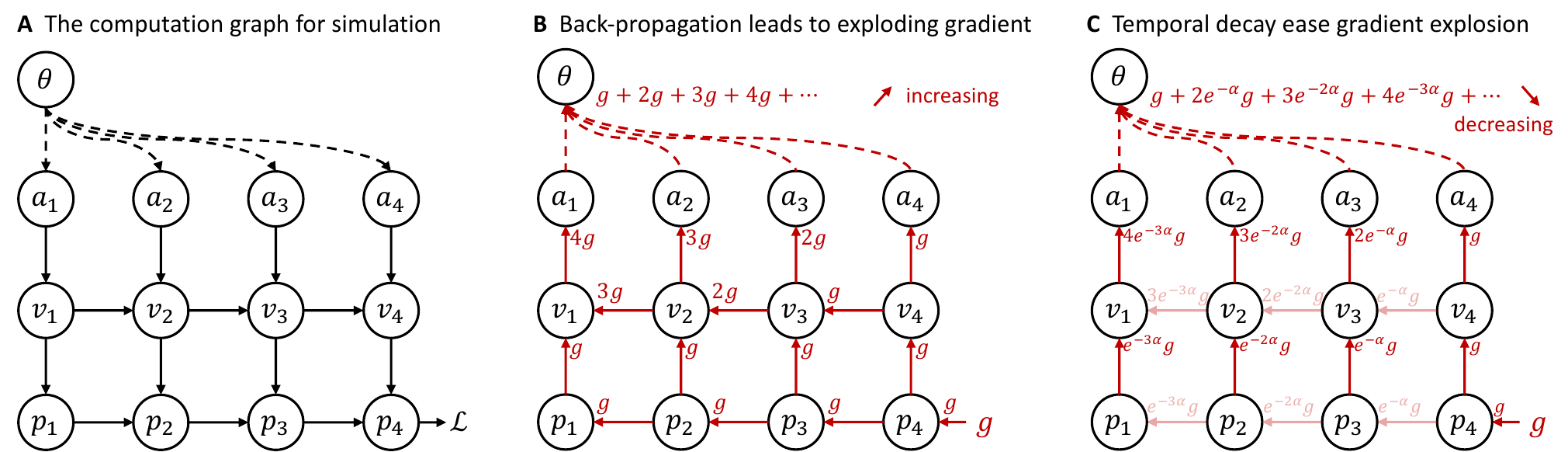}
    \caption{{\bf A computation graph of the physics simulation.} Temporal gradient decay mitigates gradient explosion. We illustrate the simplified model $v_t=v_{t-1}+a_t \Delta_t, p_t=p_{t-1}+v_t \Delta_t$.}
    \label{fig:s3}
    \label{fig:comp_graph}
\end{figure*}

\subsection{Temporal gradient decay}

Figure \ref{fig:s3} presents the computation graph for the forward simulation and backpropagation of the gradients of the point-mass physics simulation. Although the physics simulation is differentiable, direct backpropagation leads to gradient explosion \cite{bengio_learning_1994}\cite{metz_gradients_2021} when the gradient is backpropagated through the entire simulation. This results in the issue where long-range gradients from future time steps can accumulate significantly large values, resulting in unstable optimization. Intuitively, the agent can only learn to avoid relatively close obstacles, while the long-range gradients enforce the agent to avoid all future obstacles without considering the limitations of the perceptual range. 

To overcome this problem, we incorporate an exponential decay mechanism into the position and velocity gradient during backpropagation. Specifically, we multiply the gradient of $s_{t+1}$ (state at time step t+1) with respect to $s_t$ by the exponential decay factor $e^{-\alpha\Delta t}$. Here, $\alpha$ is a parameter that controls the decay rate, and $\Delta t$ is the duration of a single time step in numerical simulation. This mechanism decays gradient during backpropagation through time, producing two key effects: it mitigates gradient accumulation and ensures that the supervisory signal stems from the near future. By emphasizing immediate objectives and filtering out potential distractions from uncertain future obstacles, the adjusted gradient substantially boosts training performance.
As a result, we optimize the policy using the following gradient that has temporal gradient decay:
$\partial \mathcal{L}_{\theta} / \partial \theta = $
\begin{equation*}
  \small
       \frac{1}{N}\sum_{k=0}^{N-1} \left( \sum_{i=0}^{k} \frac{\partial l_k }{\partial x_{k}} \prod_{j=i+1}^{k} \left( \frac{\partial x_{j}}{\partial x_{j-1}} e^{-\alpha \Delta t} \right) \frac{\partial x_i}{\partial \theta} + \frac{ \partial l_k}{\partial u_k} \frac{ \partial u_k}{\partial \theta} \right) 
      .
\end{equation*}
\subsection{Network architecture}

The network utilizes a Convolutional Recurrent Neural Network (CRNN) structure \cite{shi_end--end_2016}, which combines a lightweight convolutional neural network for depth map feature extraction and a GRU \cite{cho_learning_2014} layer for consistent planning and control. \change{The depth map is inverted and max-pooled to $16\times 12$ as network input.}
Each frame passes through several CNN layers with filters (kernel size) 32(2) - 64(3) - 128(3) and then flattened and linearly projected to a 192-dim feature. All the convolution layers have a stride of 1 and are followed by a LeakyReLU activation function. Meanwhile, the target velocity, attitude estimation, and velocity estimation (optional) are linearly projected to a 192-dimensional feature and added to the corresponding image feature. The combined feature is fed into a GRU cell, and the GRU state is used to predict desired thrust acceleration and current velocity estimation through a fully connected layer.

\subsection{Hardware setup}
Our hardware setup is visualized in Figure~\ref{fig:robustness}A.
The hardware developed in this work is designed to be small, agile, and fully autonomous while remaining low-cost. We set up our quadrotor following first-person view (FPV) drone style with a total weight of only ~\SI{365}{\gram}. The aircraft is built using a Roma 3-inch frame, GEMFAN 3-inch propellers, and 1606 3750KV motors. The max thrust-to-weight ratio can achieve 3.6, thereby enabling our quadrotor's agility. We utilize a commonly adopted FPV flight tower, consisting of an Aocoda F7mini flight controller and a HAKRC 4-in-1 electronic speed controller. Furthermore, we utilize a customized BetaFlight \cite{noauthor_betaflight_2023} firmware for autonomous flight. Regarding the perception system, a RealSense D435i camera is used as a depth perception module which outputs depth images at a rate of~\SI{15}{\hertz}. 
Additionally, we use a compact \textit{Single-board computer}~(SBC) Mango Pi \cite{noauthor_mango_2023} as our onboard computer, which measures 30x65x6mm, weighs just 11g, and is equipped with a Cortex-A53 CPU. We integrated the computer into the middle of the aircraft carbon boards and combined it with the flight controller to form a ``3-layer flight tower''. Thanks to the extremely lightweight network design, our whole system can run in real time with such a compact structure.

\subsection{Related works}

Traditional approaches~\cite{Scaramuzza2014vision, liu_search-based_2018, zhou2019fast} for vision-based navigation are largely mapping-based. 
They first build a high-quality map using occupancy grids \cite{liu_search-based_2018} or the Euclidean Signed Distance Field (ESDF) \cite{zhou2019fast} as a representation of the 3D environment. 
Then, the navigation problem is converted into planning and control.
Planning typically relies on probabilistic search or gradient-based trajectory optimization, while control often utilizes model-based techniques, such as model predictive control~\cite{neunert2016fast, falanga2018pampc}. 
To enhance safety and computational efficiency, certain strategies integrate the Safe Flight Corridor (SFC) as an intermediate representation of the obstacles for planning~\cite{ji2021mapless, gao2018flying, tordesillas2021mader}.
However, dividing the navigation task into mapping-planning-and-control results in pipelines that may overlook the interplay between different stages, leading to compounded errors. Furthermore, the sequential nature of these pipelines introduces extra latency and computational burden, thereby hindering their feasibility in complex or dynamic environments. 

Different from mapping-based methods, learning an end-to-end policy for vision-based navigation can directly control the vehicle from raw sensory observations without explicitly mapping and planning.
Particularly, imitation learning is an effective approach for training a deep sensorimotor policy, in which the policy is trained by imitating either demonstration from human experts or trajectory rollouts by a model-based controller.
Imitation learning has been shown to be effective in solving several challenging tasks, including vision-based drone acrobatics\cite{kaufmann_deep_2020}, navigation in the wild \cite{vorbach2021causal, giusti2015machine, loquercio_learning_2021}, and drone racing \cite{li2018oil,kaufmann2018deep,loquercio2019deep,wang2021robust,fu2022learning}.
Although the imitation learning framework can effectively solve a specific task, it often relies heavily on the expert demonstrations. This dependency restricts the system’s generalizability beyond the initial training data. Additionally, experts are
typically tailored for a specific task, making it challenging to solve new tasks.

Differentiable physics simulation is a powerful technique that applies gradient-based methods to the learning and control of physical systems~\cite{liang2020differentiable}. 
Previous studies have demonstrated the efficiency of differentiable programming approaches in various robotics simulation tasks, such as control of soft robots~\cite{hu_difftaichi_2019, hu2019chainqueen, bern2020soft}, mimicking human motions in humanoid robots~\cite{ren2023diffmimic}, system identification~\cite{gradsim}, and quadruped locomotion~\cite{song2024learning}. 
Nevertheless, differentiable programming comes with its own challenges, most notably susceptibility to gradient explosions. Consequently, developing a practical differentiable simulation framework for tackling real-world quadrotor navigation tasks remains an open research challenge. Notably, existing autonomous quadrotor applications have yet to adopt differentiable simulation-based methods for real-world navigation.

\section{Results}

\begin{figure*}[!htp]
    \centering
    \includegraphics[width=\linewidth]{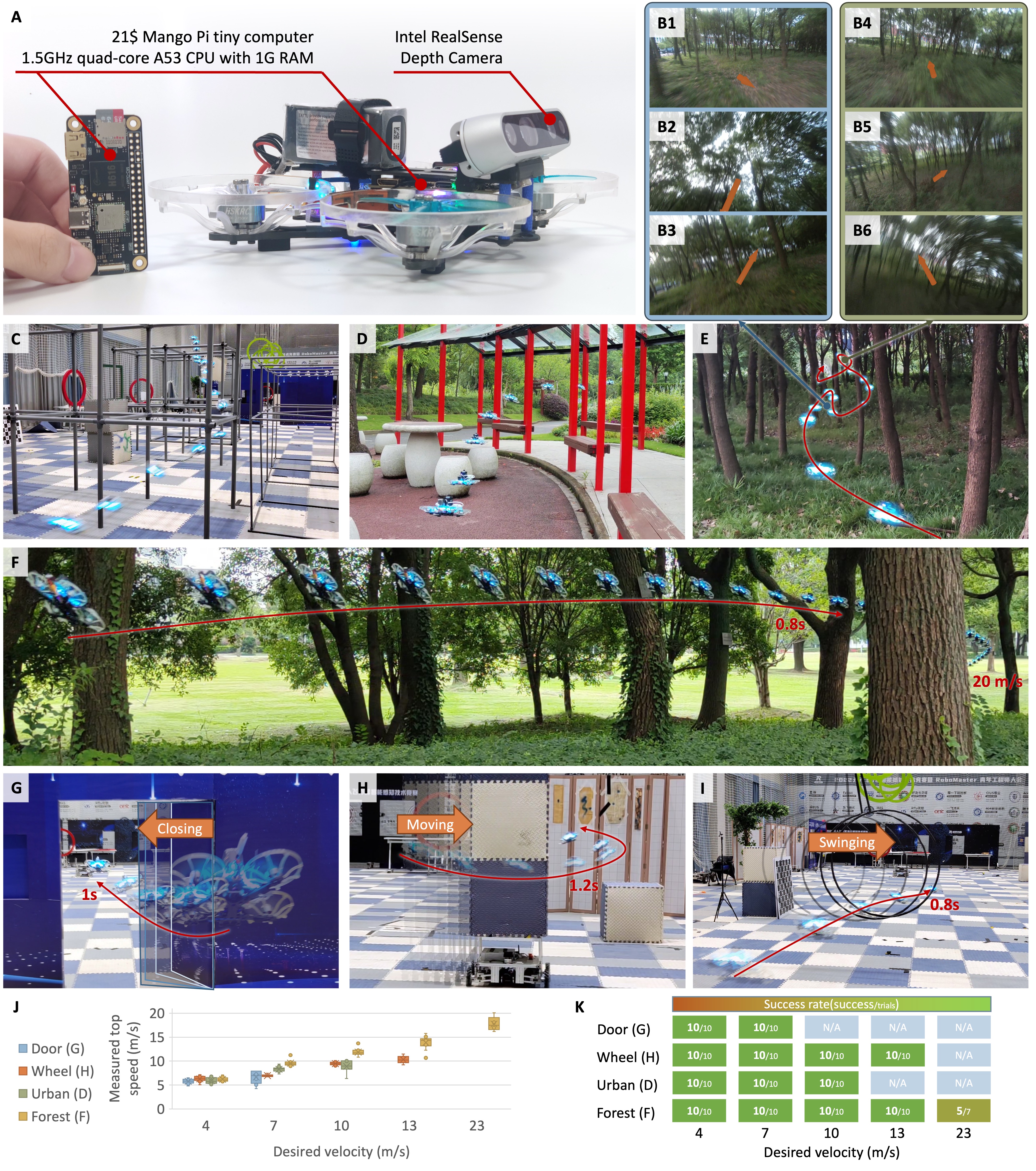}
    \vspace{-4ex}
    \caption{{\bf High-speed flight through complex and dynamic environments using a \$21 low-cost tiny computer.} (A) Experimental platform. We demonstrate the robustness of navigating in dense forests (B, E) and urban environments (C, D). Our framework can easily handle high flight speed (F) and dynamic obstacles (G-I). Flight speed and success rates (with number of success trails labeled on bars) are in (J,K). All these experiments involve end-to-end navigation without reliance on visual-inertial odometry or GPS localization \danping{for state estimation}.}
    \label{fig:robustness}
\end{figure*}

Video S1 summarizes the results and the method of this work.
In this section, we discuss the results in detail, highlighting the robustness, flexibility, and generalization capabilities of our method.
In summary, our results contain high-speed flight through complex and dynamic environments, agile communication-free swarm navigation, odometry-free flight, and benchmark comparison against state-of-the-art methods. 

\subsection{High-speed flight through complex and dynamic environments}
\label{sec:high_speed}

In this section, we demonstrated the robust navigation performance of our end-to-end policy in complex and dynamic environments. 
We train our vision-based policy using a customized simulator that incorporates differentiable dynamics for simulating the drone model and a rendering engine for rendering cluttered environments. 
After training in simulation, we deploy the resulting navigation policy directly in the physical world.
To push the limit of vision-based agile flight in the real world, we have built a simple lightweight drone that
features a low-cost tiny computer for all computational tasks onboard and an Intel RealSense depth camera for perceiving the environment~(see Fig.~\ref{fig:robustness}A).
As a result, our system has a total weight of only \SI{0.365}{\kilo\gram} and can generate a maximum thrust of \SI{13,7}{\newton}. 

Our physical system and the real-world results are summarized in Fig.~\ref{fig:robustness}. We test our system in various real-world environments, including dense forests, city parks, and indoor environments with both static and dynamic obstacles. 
We design navigation tasks by specifying an initial position and an unknown environment. 
The drone is required to navigate through the environments at high speed while avoiding both static and dynamic obstacles. 
None of the testing environments have been seen by our navigation policy during training. 
We utilize a GPS module to measure outdoor flight speed and a motion capture system for indoor environments. The actual flight speeds and the success rates at different desired speeds are shown in Fig.~\ref{fig:robustness}J,K.
The lack of higher-speed data in environments D, G, and H is attributed to the site's size limitations, which prevent drones from achieving higher speeds.
The flight recordings are included in the supplementary video.

The real-world results indicate that our end-to-end navigation policy generalizes well to real-world environments,  and exhibits excellent obstacle avoidance performance using extremely limited sensory and computational resources. Fig.~\ref{fig:robustness}B shows the flight through a dense forest that has low bushes, where high-speed traversal is very difficult due to the low visibility and the complex distribution of fine branches. Fig.~\ref{fig:robustness}C, D demonstrates our system can fly through dense obstacles at a speed of 7 m/s with a high success rate. Fig.~\ref{fig:robustness}F showcased a high-speed flight of 20 m/s in a forest. Fig.~\ref{fig:robustness}G-I showcase the agility of our system in navigating through dynamic obstacles.
Our recurrent neural network policy can effectively extract hidden information regarding the state of our vehicle and moving obstacles, facilitating agile navigation in rapidly changing environments without the need for a separate localization or motion prediction module.
Importantly, all these experiments are achieved through policy training in simulation.

A standout feature of our system lies in its robustness and capability to generalize effectively to unknown dynamic environments. 
In real-world settings, environments are frequently subject to change, including unexpected movements or appearances of obstacles. 
Dynamic environments present significant challenges: for mapping-based methods, the system delay arises from their cascaded structure and the computational burden of map construction; for learning-based methods, the issue lies in their ability to generalize unseen environments.
Fig.~\ref{fig:robustness}G-I exemplifies our agent's adaptive capability in handling various dynamic obstacles that have different motion patterns. Fig.~\ref{fig:robustness}G highlights the quadrotor's skill in navigating through a closing gate, underscoring its competence in tackling not just dynamic but also progressively complex scenarios. Fig.~\ref{fig:robustness}H reveals that the quadrotor maintains a safe distance from the moving obstacle, ensuring secure flight. Fig.~\ref{fig:robustness}I capture the agent's swift response to a swinging obstacle, showcasing the low latency and reliability of the trained agent. 
Importantly, these capabilities were developed through training in simulated settings featuring only static obstacles, further validating the efficacy and versatility of our training methodology.

\subsection{Agile communication-free swarm navigation}
\begin{figure*}[!htp]
    \centering
    \includegraphics[width=\linewidth]{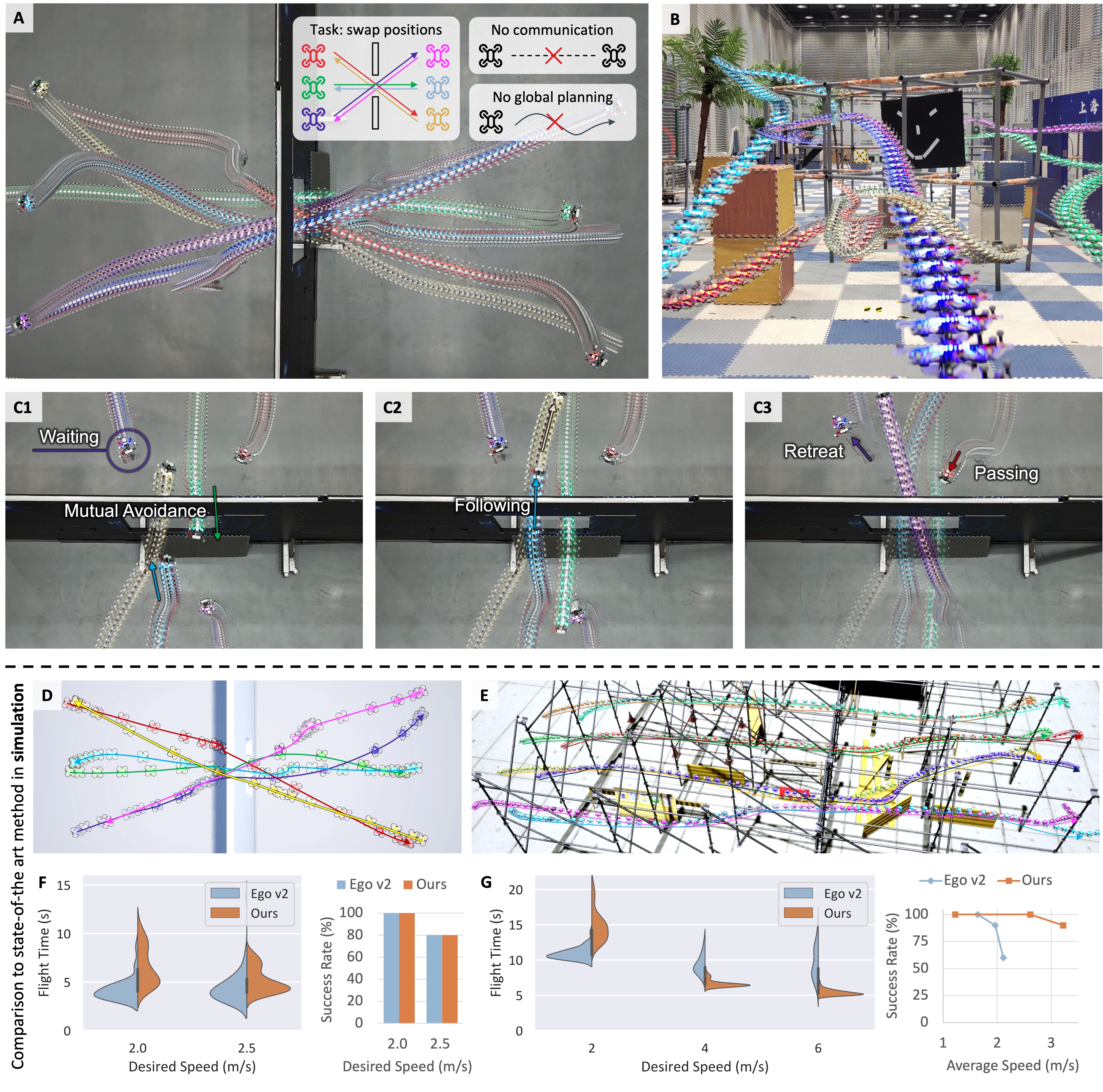}
    \caption{{\bf Communication-free swarm navigation.} 
    We perform real-world evaluation with 6 drones navigating through (A) a narrow gate in opposite directions, and (B) various obstacles in the same direction. (C) Enlarged views of individual behavior showing the emergence of collective intelligence. Video demonstrations are available. In simulated environments (D, E), the corresponding results (F, G) indicate our success rates and flight times, even without communication, are on par with Ego-v2 (which relies on communication and global positioning for mutual avoidance), while preserving our high-speed flight advantage.
    }
    \label{fig:multi_agent}
\end{figure*}

\danping{Fig.\ref{fig:results_overview}, \ref{fig:results_enlarged}, and    \ref{fig:multi_agent}B illustrate our approach's ability to navigate multiple aerial robots through cluttered environments filled with various static or moving obstacles without explicit communications between robots.
To further demonstrate the applicability of our approach to challenging swarm navigation tasks, we design a challenging task: a team of drones must swap their positions by navigating through a narrow gate in opposite directions. }

This task \danping{exemplifies how} our approach trains decentralized policies enabling drones to coordinate with each other using only local sensory information to avoid collisions with the environment and other drones.
\danping{Moreover, our training approach is inherent flexible, requiring only minimal modifications for multi-agent tasks. The training process remains the same as in the single-agent setting, except that each agent is represented as a spherical obstacle within the simulation.}
After training, the policy operates \danping{independently for each robot} using the same observation modalities, including depth measurement and orientation, as in the single-agent scenario.

Fig.~\ref{fig:multi_agent}A summarizes the result of \danping{the task of swapping position through a narrow gate.} %
We conducted real-world experiments using six quadrotors, which were divided into two groups of three agents each.
Each group is positioned on opposite sides of the testing arena. 
The objective is for all six agents to navigate through a narrow gate simultaneously while reaching a destination. 
Due to the small size of the gate, it is impossible for all six drones to pass through the gate simultaneously. 
Hence, the agents are required to coordinate.
\danping{To ensure each agent can reach its target position correctly, we employ a motion capture system to obtain reference directions.  
We use velocity estimation from the motion capture system as input to each neural network policy. Each agent is required to follow its own velocity command while without having access to the information about the state of other agents. Therefore, the agents are required to make individual decisions without any communication.}
We define a run as successful if all drones reach their respective goals within a radius of~\SI{1.2}{\meter} without any collisions.

\danping{The result shows that} all six drones successfully passed through the gate and reached their destination.
\danping{Interestingly, the trained vision-based agents} demonstrated \danping{emerging} self-organized behaviors - such as waiting in air traffic, following an agent, and \danping{retreating} to give way to incoming agents - all without the need for communication, global planning, \danping{or adopting a coordination loss in the training objective.} 

Furthermore, we conduct experiments to quantitatively compare our approach with the state-of-the-art \danping{centralized} planning method \danping{for swarm navigation} Ego v2 \cite{zhou_swarm_2022}. We evaluate the performance in terms of success rates and task completion time. 
Fig.~\ref{fig:multi_agent}F-H summarizes the results: our communication-free approach achieves comparable performance as the communication-based method. 
 Notably, existing methods heavily rely on network communication and global planning for effective coordination and mutual avoidance. 
\danping{However, communication can be unreliable in real-world scenarios, and centralized planning requires accurate relative positioning between robots, which is difficult to acquire, particularly in the absence of GPS.}
Consequentially, Ego v2 adopts ad hoc UWB positioning and drone detection in real-world experiments. 

Much like humans or other animals, our agents achieve self-organization \danping{using only} local sensing, making them \danping{ideal} for scenarios where communication is impractical. These capabilities enable 
deploying autonomous swarms in environments \danping{without} global positioning \danping{or} communication infrastructure, \danping{such as} search and rescue missions or reaching remote locations. Our approach provides a robust solution for swarm navigation and obstacle avoidance in these challenging conditions.

\subsection{Odometry-free flight}

To further evaluate the flexibility and robustness of our end-to-end \danping{flight} system, we study the role of state estimation in obstacle avoidance. State estimation is a central requirement for navigating a mobile robot. Traditional approaches to vision-based navigation predominantly depend on globally consistent state estimation. Methods of this kind rely heavily on the quality of visual-inertial odometry, which is known to be very sensitive during high-speed flight~\cite{delmerico2019we,cioffi2023learned,kaufmann_champion-level_2023}.
High-speed motion can lead to large displacements and little overlap between consecutive frames that make feature tracking difficult \cite{delmerico2019we}, resulting in the failure of the visual localization system. 
Additionally, visual localization and mapping demand considerable computational resources. 
For instance, in the Ego-v2 system~\cite{zhou_swarm_2022}, CPU usage for localization surpasses the combined usage of all other processes, even when GPU acceleration is employed.

The inherent flexibility of our approach allows for training agents \danping{without the position and velocity estimates as part of the input}. Consequentially, the resulting odometry-free agent does not require a standalone localization module to estimate the position and velocity but still learns to navigate and control using only attitude estimation and depth map.
\change{Note that our method employs a recurrent neural network that maintains an internal state estimate, enabling the system to operate without \danping{explicit} odometry for state estimation when deployed on a real drone.}
In this way, we further exploit the potential of neural networks to achieve end-to-end localization, planning, and control.
As subsequent results will demonstrate, our trained agent operates robustly even in the absence of localization input and delivers performance metrics comparable to scenarios utilizing \change{external} localization.

\begin{figure*}[]
    \centering
    \includegraphics[width=\linewidth]{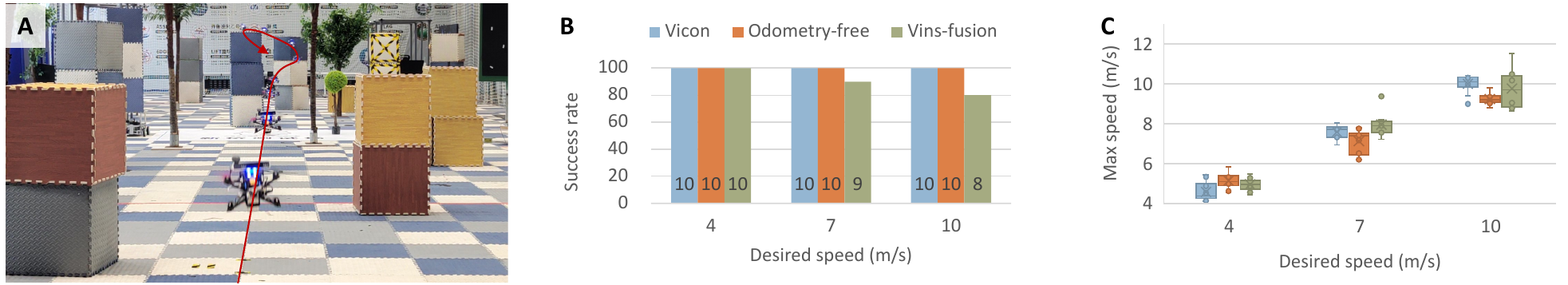}
    \caption{ {\bf End-to-end vision-based \danping{flight} without \danping{an explicit odometry module}.} (A) The controlled testing environment. Our neural network achieves reliable performance by \danping{learning implicit state estimation from} flight dynamics and previous control commands via a memory-based network. (B, C) Our \danping{odometry-free end-to-end} system demonstrates high success rates, yielding results comparable to those achieved through an external motion capture system. }
    \label{fig:loc}
\end{figure*}

We \change{assess the performance of our odometry-free agent by comparing its success rate and flight speed against our baseline agent that use external velocity estimation inputs (odometry-based) in real-world scenarios}. \change{For odometry-based agents}, we evaluate two distinct source of velocity estimation: one utilizing a VICON motion capture system and another employing a visual-inertial odometry system (VIO)~\cite{qin_vins-mono_2018}. The testing environment, as shown in Fig.~\ref{fig:loc}A, \change{includes a relatively open area at the front designed to enable the drone to achieve high speeds, along with obstacles of cubes and trees at the rear. We evaluated the drone's speed and success rate to reach the area behind the obstacles. We conducted tests at three different desired speeds: 4, 7, and \SI{10}{\meter\per\second}. Each test is repeated 10 times.}

\change{The results are shown in Figure~\ref{fig:loc}. Figure~\ref{fig:loc}B compares the success rates of our odometry-based agents against our odometry-free agent. The bars are labeled with the number of successful trials for each test. Figure~\ref{fig:loc}C the statistics of the maximum speeds achieved during the trials.
Both our odometry-free agent and the odometry-based agents utilizing VICON for state estimation successfully reaches the target without crashing. In contrast, the agent that relies on VIO for state estimation exhibits a lower success rate, especially at higher speeds. This performance is consistent with the previously mentioned limitation of VIO systems, which can suffer from drift at high speeds, resulting in failures such as collisions with obstacles like trees.
}
In contrast, our odometry-free agent outperforms the agent that rely on the VIO system, achieving the same level of performance as the agent that uses the VICON external motion capture.

It is important to point out that the majority of current planning and control algorithms still depend on explicit \danping{state estimation}~\cite{delmerico2019we}. 
\danping{The experiment suggests} that standalone state estimation modules might not be necessary for \danping{autonomous flight. This finding has been consistently validated by our results} \change{in Sec.~\ref{sec:high_speed}, where we show the effectiveness of odometry-free flight in various challenging scenarios in the real world.}

\subsection{Baseline comparisons}

We carry out comprehensive experiments in simulation to benchmark our approach against existing methods, including a traditional mapping-based method~\cite{zhou_swarm_2022} and a learning-based method~\cite{loquercio_learning_2021}. 
To ensure a fair comparison, we utilized the Flightmare simulator~\cite{song_flightmare_2021}, following the experimental setup outlined in~\cite{loquercio_learning_2021}.
Additionally, we use the AirSim simulator~\cite{shah_airsim_2017} for evaluation in other additional environments.
We tested all methods with different target speeds—4, 7, 10, 13, and 22~\SI{}{\meter\per\second}. 
We repeat each experiment 10 times and compute the average flight speed, peak flight speed, and success rate. A task is considered to be successful if the drone reaches its goal position within a 5-meter radius without crashing. 
Average speeds were calculated based on these successful trials.

\begin{figure*}
    \centering
    \vspace{-3em}
    \includegraphics[width=\linewidth]{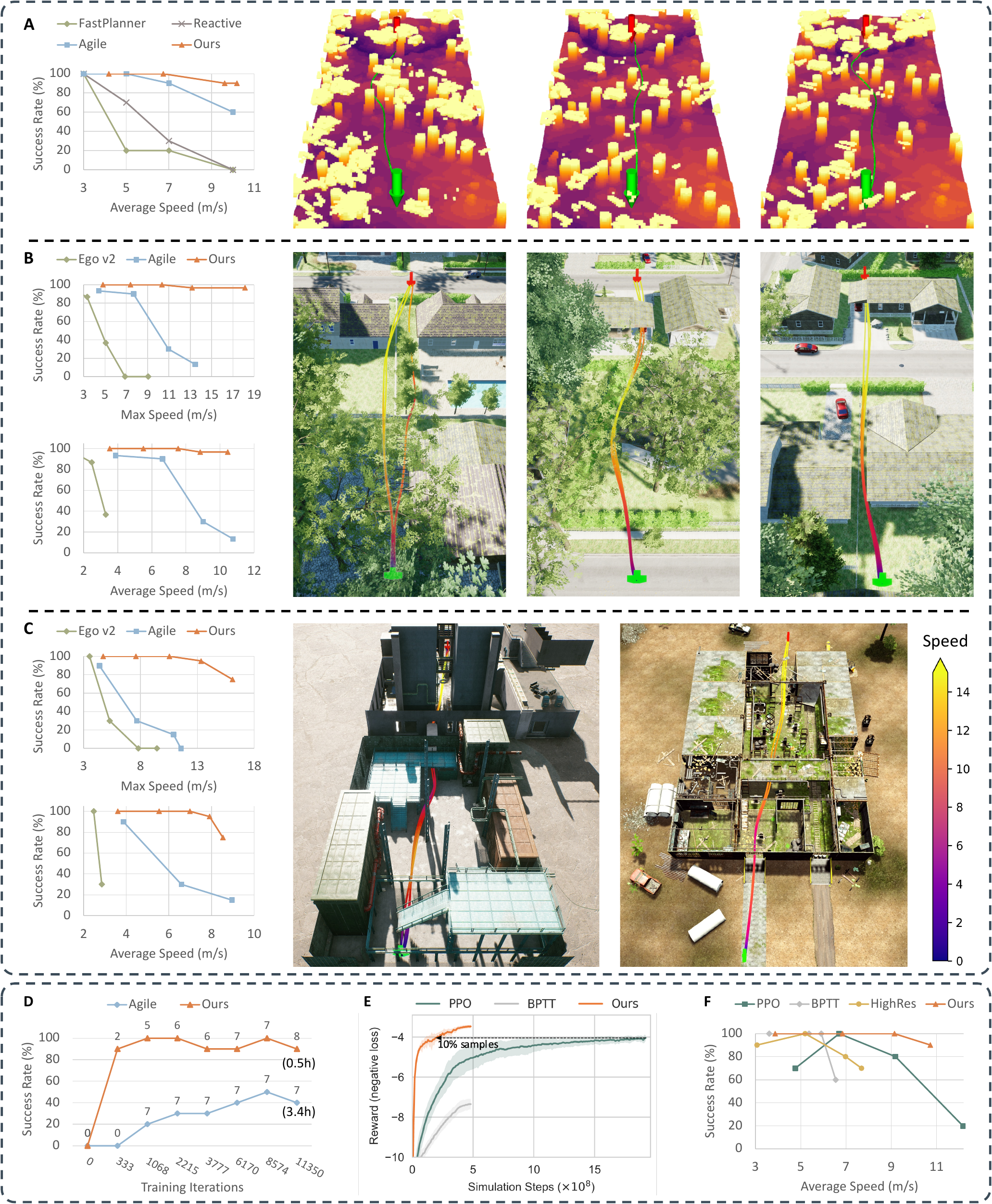}
    \caption{ 
    {\bf Baseline comparisons for vision-based agile flight in cluttered environments. }
    (A-C) Comparing our method against state-of-the-art vision-based navigation methods. Our method shows the best success rates in all tested environments. 
    (D) Our approach outperforms a state-of-the-art imitation learning method with better success rate and less training time. 
    (E) Comparison among our approach, reinforcement learning, and vanilla differentiable physics (BPTT) in policy training.
    (F) Robustness study.
    }
    \label{fig:robustness_sim}
\end{figure*}

The results are summarized in Figure~\ref{fig:robustness_sim}.
The traditional mapping-based method~\cite{zhou_swarm_2022} (termed Ego v2) performs relatively well at very low speed, e.g., below~\SI{3}{\meter\per\second}; however, it experiences significant performance drops as the target speed increases. 
This aligns with the real-world experiments demonstrated in the original paper~\cite{zhou_ego-swarm_2021}, in which the average flight speed is around~\SI{2}{\meter\per\second}.
Our findings also confirm the results reported in~\cite{loquercio_learning_2021}, where they attribute such failures to the increased processing and control latency inherent in cascaded structures.

While the learning-based method~\cite{loquercio_learning_2021} (called Agile) performs well in the forested environment, this specific environment was also used for training their navigation policy. 
When testing in an unseen environment, e.g., in an urban setting Fig.~\ref{fig:robustness_sim} B-C, \cite{loquercio_learning_2021} achieves very low success rates. 
Our results indicate that expert-based approaches struggle with generalization to unseen environments, hence, resulting in significant performance drops.

Such shortcoming is primarily due to the difficulty in obtaining high-quality expert labels in obstacle-dense, high-speed conditions~\cite{loquercio_learning_2021}.
Additionally, the sampling-based expert itself faces challenges in consistently identifying the optimal route, leading to erratic flight patterns in the trained agents, characterized by large lateral oscillations in their real-world flight. 
The incorporation of error corrections through the DAGGER framework~\cite{ross_reduction_2011}, along with frequent trajectory replanning, exacerbates the drone's unstable behavior further.
Lastly, the expert-based approach tends to maintain a constant speed, lacking the flexibility to adapt flight speed in response to varying obstacle densities.

In contrast, our approach consistently achieves high success rates, even at high speeds. Our framework seamlessly integrates planning optimization with neural network training, eliminating the need for hard-to-design expert planners. This integration empowers our method to effectively learn and adapt to complex environments. Furthermore, the end-to-end design of the sensorimotor agent enhances coordination between planning and control, achieving more stable control during replanning and automatic speed adjustments in response to varying obstacle densities.
We verify that our framework surpasses the expert-based method, specifically Agile \cite{loquercio_learning_2021}, in terms of convergence speed. In our experiment, we extract the point cloud from training scenes used by Agile and evaluated success rates and average speeds at various optimization steps. Fig.~\ref{fig:robustness_sim}D shows the success rate for different training iterations at~\SI{7}{\meter\per\second}, with the average speed and training time annotated on the data point. The result shows the faster convergence of our method: by the $\rm 1k^{th}$ iteration, our approach has already achieved a 100\% success rate at an average speed of 5 m/s, while Agile \cite{loquercio_learning_2021} is only starting to show success, with a mere 20\% success rate.

Additionally, we established a reinforcement learning baseline with the Proximal Policy Optimization (PPO) algorithm ~\cite{schulman_proximal_2017}.
Fig~\ref{fig:robustness_sim}E shows a comparison of the learning curve---our differentiable physics quickly achieves high rewards given limited simulation steps, while RL requires substantially more samples.
\change{Our method requires only 10\% samples of the PPO algorithm to achieve its maximum reward level.}
We attribute this improved convergence to the lower variance of first-order gradients inherent in the smooth and well-behaved point mass model\cite{suh2022differentiable}.
Fig.\ref{fig:robustness_sim}F presents a validation of the trained policy in the same forest environment, where our approach achieves higher success rates across all tested target velocities.

\danping{Although leveraging differentiable physics is the key to our approach, we found that directly applying the vanilla differentiable physics method did not work properly, primarily due to gradient explosion, as pointed out in  \cite{metz_gradients_2021}. To address this issue, we designed a temporal gradient decay mechanism for training, as described in the method section (see Fig.\ref{fig:comp_graph}).}
We validate this design and the use of low-resolution depth maps using the Flightmare simulator \cite{song_flightmare_2021} within the forest scene. For all methods, we follow \cite{loquercio_learning_2021} and use the depth map from SGM as the input. In Fig.~\ref{fig:robustness_sim}, ``BPTT'' stands for not applying gradient decay ($\alpha=0$). For high-resolution (HighRes) experiments, we train and test on $128 \times 96$ depth images with the MobileNetV3 \cite{howard_searching_2019} backbone.

The results in Fig. 6E, F indicate the importance of having temporal gradient decay for calculating the policy gradient during policy training.  
We show that training with vanilla differentiable physics (BPTT) has slower convergence and achieves sub-optimal performance. 
As a result, the BPTT agent agent exhibits a notably lower overall success rate and fails to achieve higher speeds.
Finally, while using higher-resolution depth maps improves performance during training, it results in lower success rates and flight speeds during testing. The reason may be that training with a higher input resolution can lead to overfitting to environmental features and demands higher quality depth maps at the inference stage. However, the real-world depth maps are usually noisy. Using a lower-resolution depth map aids in bridging the gap between simulation and reality.

\section{Discussion}

This work demonstrated the effectiveness of differentiable physics in agile vision-based navigation. 
Contrary to reinforcement learning and imitation learning, which often treat the robot's physics model as a black box, our approach leverages differentiable physics simulation to train a neural network flight policy. This enables end-to-end optimization of the neural network controller by backpropagating loss gradients directly through the robot's physics model. As a result, our methodology eliminates the need for task-specific expert labels, as required in imitation learning. 
Compared to reinforcement learning methods, which rely on zeroth-order optimization and consequently require a large number of samples, differentiable physics utilize gradient information for first-order optimization, leading to a significant improvement in sample efficiency.

We showcased the effectiveness of our method in vision-based flight in real-world scenarios. Our flight policy achieved zero-shot generalization to the physical world, excelling in flight speed and robustness across various complex environments with both static and dynamic obstacles.
Furthermore, our framework demonstrates remarkable scalability in multi-agent settings. Tests with a swarm of quadrotors showed agile navigation through dynamically changing environments and confined spaces, even without explicit communication or centralized planning algorithms. Remarkably, our policy displayed coordinated behaviors, like waiting and following, without explicit communication.
Moreover, our flight policy is effective in partially observable settings without requiring the vehicle's full state. This capability not only simplifies system design but also provides immense benefits for deployment in unknown or unpredictable settings.

Regarding computational efficiency, we validated our system with extremely limited sensory and computational resources. 
We have implemented our system on a budget-friendly \$21 microcomputer without relying on GPS or visual-inertial state estimators for localization. This achievement is particularly noteworthy as it demonstrates the minimal computational and sensory resources required by our approach, making it highly accessible and deployable in resource-constrained settings.
Despite these computational constraints, our system outperforms state-of-the-art systems~\cite{zhou_swarm_2022, loquercio_learning_2021} that rely on GPU-based computations.

Looking ahead, there are several avenues for future research. Our framework's adaptability in task design---spanning network input, output, loss functions, and environmental settings---has already proven effective in localization-free obstacle avoidance and swarm navigation without communication. Such versatility paves the way for a broader range of applications in fields such as search and rescue, logistics, inspection, entertainment, and agriculture.

\section*{Acknowledgment}

We thank SJTU SEIEE · G60 Yun Zhi AI Innovation and Application Research Center for indoor experiment support, J. Li for initial efforts in the multi-agent experiments, and L. Zhang, F. Yu for helping with the experiments and the valuable discussions.

\appendices
\renewcommand*{\thefigure}{S\arabic{figure}}
\renewcommand*{\thetable}{S\arabic{table}}
\renewcommand*{\thesection}{S\arabic{section}}
\setcounter{figure}{0}
\setcounter{table}{0}
\setcounter{section}{0}

\section{Calibration and system identification}

\begin{figure*}
    \centering
    \includegraphics[width=\linewidth]{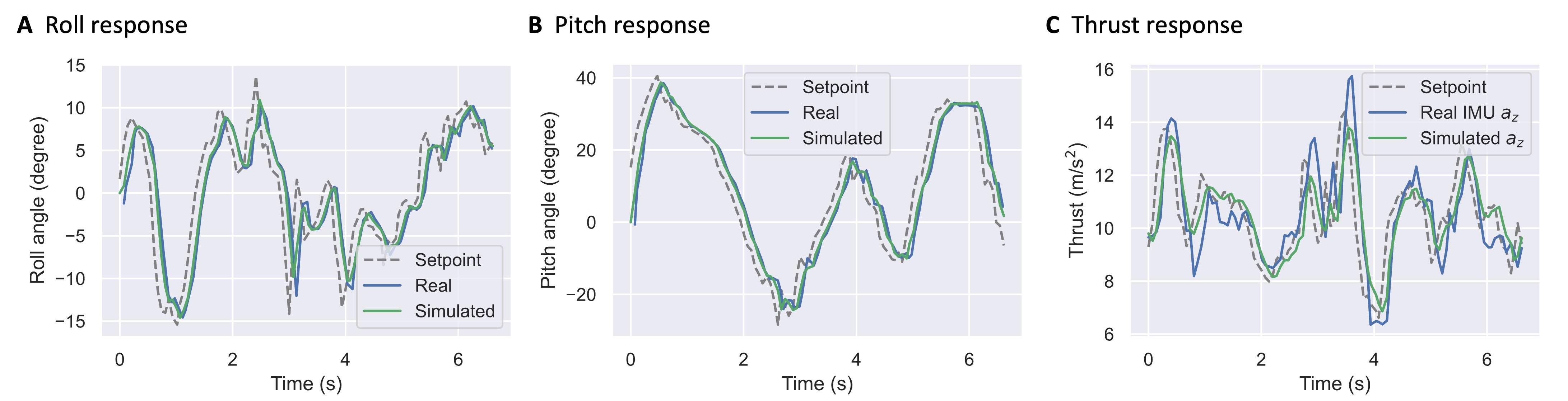}
    \caption{{\bf Flight control response in simulation.} The simulation closely mirrors our real drone, demonstrating the accuracy of our calibration.}
    \label{fig:s1}
\end{figure*}

For the effective training of our quadrotor navigation framework, it's pivotal that our simulations mimic real-world flight dynamics accurately. The primary goal of calibration is to identify the parameters of our differentiable physics simulation, ensuring that simulated dynamics mirror real-world quadrotor responses. Calibration involves adjusting for control latency and air drag.

\subsection{Control latency simulation}

Our simulation considers a fixed control latency and an exponential moving average for modeling the flight controller response. We fix the proportional attitude controller with an attitude control gain of 13. We perform calibration experiments on roll and pitch response to measure the latency parameters $\lambda, \tau$, yielding an estimate of $\lambda=12, \tau=1/15$. Fig.~\ref{fig:s1} A and B shows the roll and pitch response from a real-world flight, where the simulated responses closely match the real-world responses.

Measuring thrust in the real world is more challenging because the acceleration measurements are mixed with gravity, air drag, and non-linear thrust model. Nevertheless, we have verified that the delay parameters for roll and pitch are applicable for thrust as well. Results from Fig.~\ref{fig:s1}C shows that our simulation aligns closely with the real-world thrust delay.

We have verified the consistency of this latency across several platforms, provided the attitude control gain remains uniform: a) our platform shown in Fig.~3A using BetaFlight; b) a larger~\SI{0.6}{\kilogram} drone with 4-inch propellers also using BetaFlight; c) a~\SI{1.2}{\kilogram} drone with 5-inch propellers running a PX4 controller; d) the AirSim \cite{shah_airsim_2017} simulator; e) the Flightmare \cite{song_flightmare_2021} simulator.

\subsection{Air drag identification}

\begin{figure*}
    \centering
    \includegraphics[width=\linewidth]{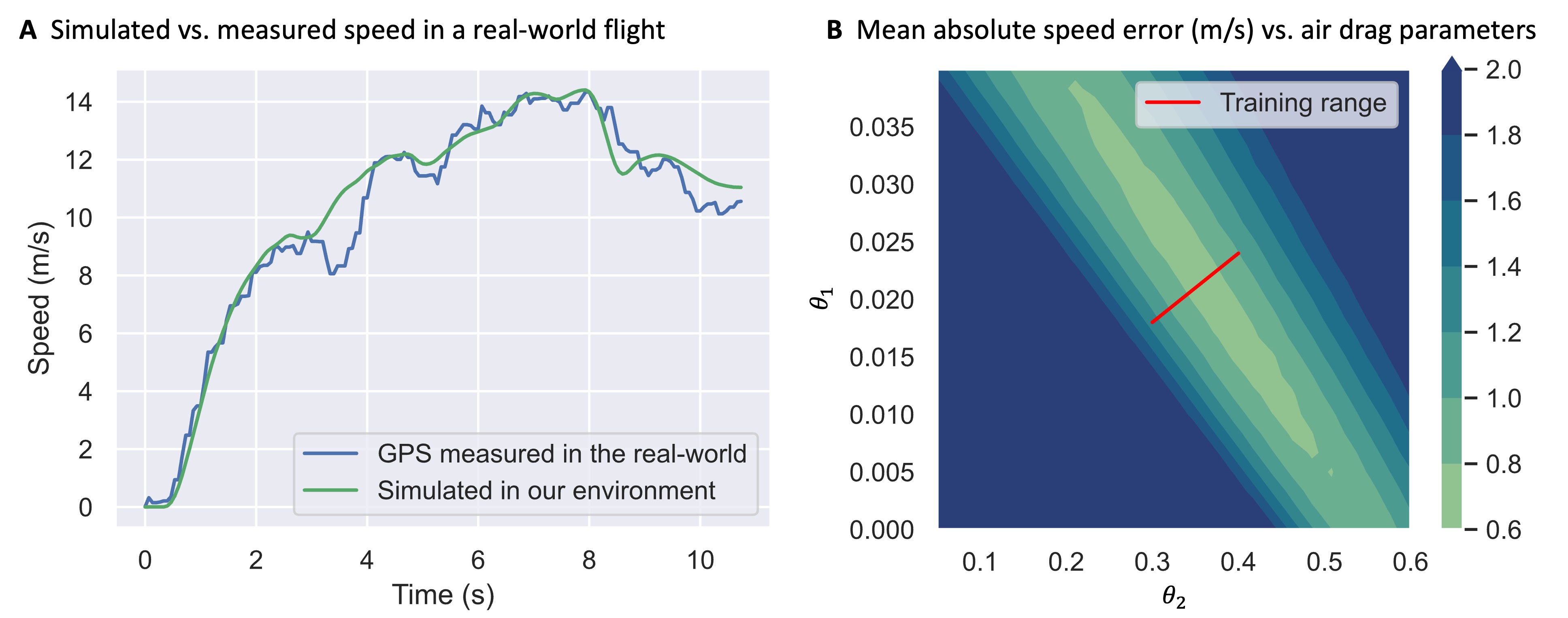}
    \caption{{\bf The simulated flight velocity v.s. our real drone.} The relatively small difference between the simulation and the real drone shows the fidelity of our air drag simulation.}
    \label{fig:s2}
\end{figure*}

In low-speed navigation, air drag is frequently overlooked. However, for high-speed sensorimotor flights, it's a critical factor. At speeds exceeding~\SI{9}{\meter\per\second}, our 3-inch quadrotor (refer to Fig. 3A) with propeller guards encounters considerable air resistance, measuring over~\SI{5}{\meter\per\second^2}. With fixed agent actions, air drag largely dictates the flight speed. For localization-free flights, air drag plays a primary role in the implicit estimation of velocity.

Air drag varies depending on the platform. To calibrate it, we've employed a straightforward method. Initially, we record actions from real-world flights and simulate using the drag model $a_{drag}=-\theta_1\|v\|v-\theta_2v$, where $\theta_1$ and $\theta_2$ are undetermined parameters. Then, we conduct a grid search for the parameters and compare the simulated speed to the GPS recorded speed (as shown in Fig.~\ref{fig:s2}A). The parameter set that yields the smallest speed discrepancy is adopted as the drag coefficients. To enhance model robustness, we train using a diverse set of drag coefficients, centered around the calibrated values (as depicted in Fig.~\ref{fig:s2}B).

\section{Temporal gradient decay}

Temporal gradient decay, as discussed in Section B, is an essential technique to ensure stability and efficiency in training, especially for long-range sequences. This method manages the accumulation of gradients backpropagated through time, ensuring that the significance of distant obstacles vanishes over time. We first highlight the challenge posed by gradient accumulation (refer to Fig.~\ref{fig:comp_graph}A, B) and subsequently demonstrate how temporal gradient decay addresses the issue of exploding gradients (see Fig.~\ref{fig:comp_graph}C). Ultimately, we illustrate how the decayed gradient teaches the agents to maneuver at the optimal moment.

Fig.~\ref{fig:comp_graph} shows the graphical model of simulation spanning 4 timesteps, with the objective function being evaluated based on the position at t=4. For simplicity, we take $\Delta_t=1$ without loss of generality. Fig.~\ref{fig:comp_graph}B shows the backward computation. Here, the positional gradient, denoted as g, is backpropagated through the entire computation graph. The numerical integration structure further leads to a growing gradient signal that ultimately accumulates to the model parameters. In contrast, when gradient decay is applied during temporal backpropagation, the gradients affecting the parameters result from the product of a linearly increasing term and an exponentially decaying term.

\begin{figure}
    \centering
    \includegraphics[width=3.5in]{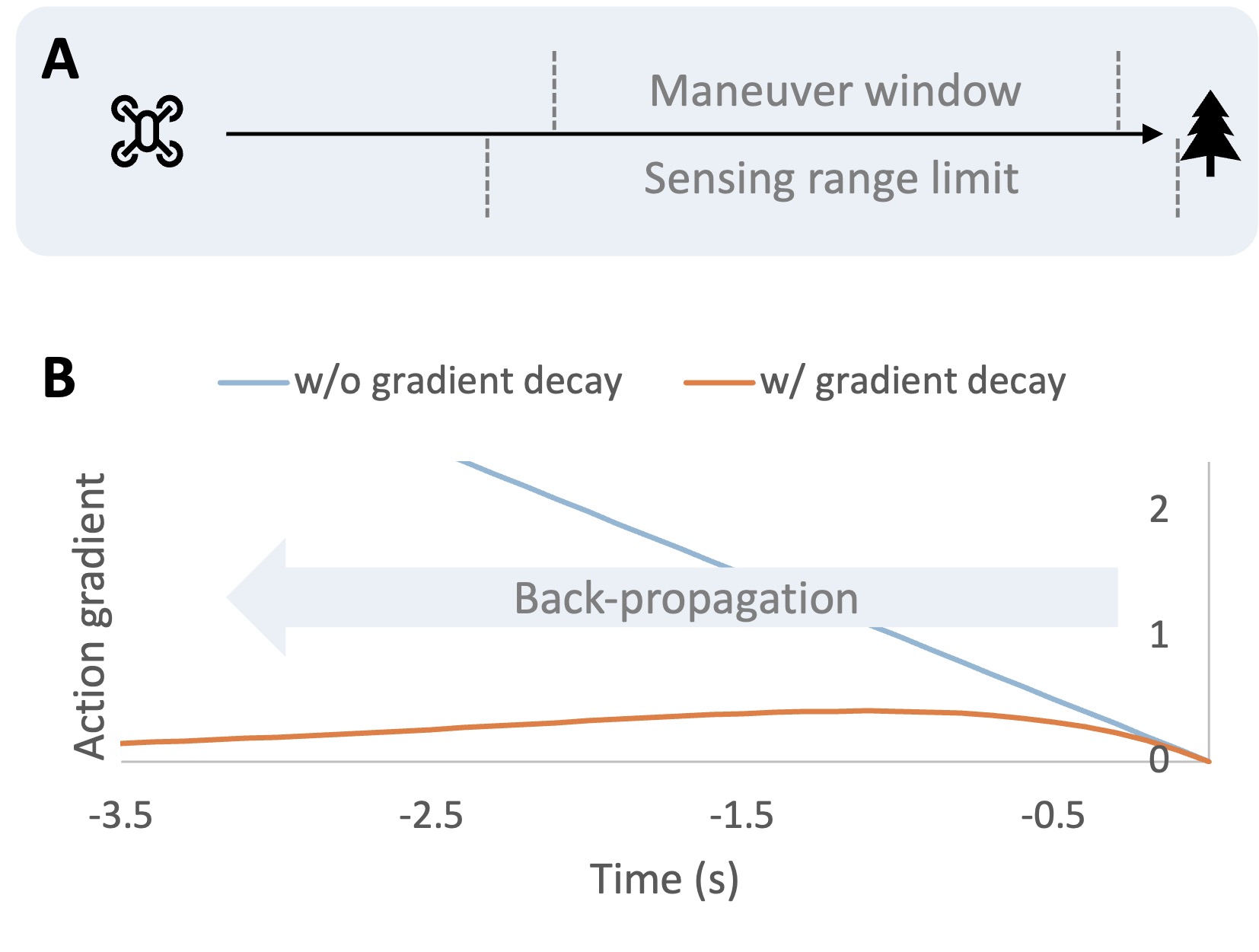}
    \caption{{\bf Temporal gradient decay aligns the supervisory signals with the sensing range.} We show a scenario (A) where the agent approaches an obstacle. Graph (B) shows how gradient decay aligns training signals within the sensing range.}
    \label{fig:s4}
\end{figure}

Fig.~\ref{fig:s4} shows the advantage of temporal gradient decay in aligning supervisory signals with the quadrotor's sensing range. The blue line in Fig.~\ref{fig:s4}, representing the gradient without temporal gradient decay, displays an ever-increasing gradient. This misleads the agent with the impractical task of avoiding obstacles beyond its sensing range. The gradient, after the application of decay, initially rises and then diminishes to zero as it's backpropagated over time. This dynamic achieves a balance, addressing the inherent conflict that earlier maneuvers provide more significant benefits for obstacle avoidance, yet distant obstacles are often more challenging to perceive. Thus, the agent is directed to concentrate more on proximate obstacles that are immediate and more perceptible.

By prioritizing immediate and perceptible obstacles, the training process can converge faster as the training objective becomes more feasible, and the agent receives more relevant feedback. This method ultimately foresters a more robust navigation system.

\section{Training hyperparameters}

The training hyperparameters is listed in Table~\ref{table:s1}.

\begin{table}[!ht]
    \centering
    \begin{tabular}{l|c}
        \toprule
        \textbf{Training Parameter} & \textbf{Value} \\ 
        \midrule
        optimizer & AdamW \\
        learning rate & 0.001 \\
        learning rate schedule & cosine decay \\
        weight decay & 0.01 \\
        batch size & 64 \\
        number of time steps $N$ & 150 \\
        $\Delta t$ & 1/15~\SI{}{\second} \\
        gradient decay rate $\alpha$ & 0.92 \\
        training iterations & 50000 \\
        \bottomrule
    \end{tabular}
    \caption{{\bf The hyperparameters used for training.} Notably, we add a small perturbation on the simulation time step $\Delta t$ during training. }
    \label{table:s1}
\end{table}

\section{Training details of the reinforcement learning baseline}

This section provides training details for the reinforcement learning baseline. Specifically, we use a state-of-the-art model-free reinforcement learning algorithm called Proximal Policy Optimization (PPO)~\cite{schulman_proximal_2017}. 
We use parallelized simulation for data collection, in which the number of environments is identical to that of our policy training using differentiable physics. 
The PPO hyperparameters are given in Table~\ref{table:s2}. 

\begin{table}[!ht]
    \centering
    \begin{tabular}{l|c}
        \toprule
        \textbf{Training Parameter} & \textbf{Value} \\ 
        \midrule
        optimizer & AdamW \\
        weight decay & 0.01 \\
        discount factor & 0.99 \\
        GAE $\lambda$ & 0.95 \\
        number of time steps $N$ & 150 \\
        $\Delta t$ & 1/15~\SI{}{\second} \\
        clip range & 0.2 \\
        entropy coefficient & 0.001 \\
        parallel environments & 256 \\
        training iterations & 12,500 \\
        minibatch size & 38,400 \\
        mini epochs & 5 \\
        \bottomrule
    \end{tabular}
    \caption{{\bf Hyperparameters for the Proximal Policy Optimization (PPO) Algorithm.}}
    \label{table:s2}
\end{table}

\balance
\bibliographystyle{IEEEtran}
\bibliography{scibib}

\end{document}